\documentclass[remotesensing,article,accept,pdftex,moreauthors]{Definitions/mdpi} 
\preto{\abstractkeywords}{\nolinenumbers} 
\usepackage[labelformat=simple]{subfig}
\usepackage{graphicx}
\usepackage{pifont}       % \ding{xx}
\usepackage{bbding}       % \Checkmark,\XSolid,... (需要和pifont宏包共同使用)
\usepackage{fontawesome}  % \faCheck,\faTimes
\usepackage{xcolor}

\firstpage{1} 
\pubvolume{16}
\issuenum{11}
\articlenumber{2024}
\pubyear{2024}
\copyrightyear{2024}
\externaleditor{Academic Editor: Richard Gloaguen}
\datereceived{25 March 2024} 
\daterevised{9 May 2024} % Comment out if no revised date
\dateaccepted{3 June 2024} 
\datepublished{5 June 2024} 
\hreflink{https://doi.org/\linebreak10.3390/rs16112024} % If needed use \linebreak
\Title{\textls[-15]{Two-Stage Adaptive Network for Semi-Supervised Cross-Domain Crater Detection under Varying Scenario Distributions}}

\TitleCitation{Two-Stage Adaptive Network for Semi-Supervised Cross-Domain Crater Detection under Varying Scenario Distributions}

\Author{Yifan Liu %MDPI: Please carefully check the accuracy of names and affiliations.
 $^{1}$\orcidA{}, Tiecheng Song %MDPI: Dear Authors, (1) Please proofread the article format according to the comments left and pay attention to the highlighted parts. (2) Please do not delete the comments during your proofreading, since we need to check point by point after receiving your proofed version. (3) Please just modify or confirm or give feedback with track change Function. (4) English editing and layout have been complete, please do not revert the changes.
 $^{2,}$*%MDPI: Added correspondence symbol according to the submitting system, please confirm
 \orcidB{}, Chengye Xian $^{1}$, Ruiyuan Chen $^{2}$, Yi Zhao $^{1}$\orcidC{},  Rui Li $^{3}$ and Tan Guo $^{2}$}

\AuthorNames{Yifan Liu, Tiecheng Song, Chengye Xian, Ruiyuan Chen, Yi Zhao, Rui Li and Tan Guo}

\AuthorCitation{Liu, Y.; Song, T.; Xian, C.; Chen, R.; Zhao, Y.; Li, R.; Guo, T.}
\address{%
$^{1}$ \quad School of Computer Science and Technology, Chongqing University of Posts and Telecommunications, Chongqing 400065, China;  2021212488@stu.cqupt.edu.cn (Y.L.); 2020210689@stu.cqupt.edu.cn (C.X.); 2021212252@stu.cqupt.edu.cn (Y.Z.)\\
$^{2}$ \quad \textls[-25]{School of Communications and Information Engineering, Chongqing University of Posts and Telecommunications, Chongqing 400065, China; 2021210292@stu.cqupt.edu.cn (R.C.); \mbox{guot@cqupt.edu.cn (T.G.)}} \\
$^{3}$ \quad  {International College of} %MDPI: We suggest you split this address with comma, please check.
 Chongqing University of Posts and Telecommunications, Chongqing 400065, China; 2021215074@stu.cqupt.edu.cn  }

\corres{Correspondence: songtc@cqupt.edu.cn}

\abstract{
Crater detection can provide valuable information for {humans} to explore the topography and understand the history of extraterrestrial planets. Due to the significantly varying scenario distributions, existing detection models trained on known labelled crater datasets are hardly {effective when} applied to new unlabelled planets. To address this issue, we propose a two-stage adaptive network (TAN) for semi-supervised cross-domain crater detection. Our network is built on the YOLOv5 {detector,} where a series of strategies are employed to enhance its cross-domain generalisation ability. In the first stage, we propose an attention-based scale-adaptive fusion (ASAF) strategy to handle objects with significant scale variances. Furthermore, we propose a smoothing hard example mining (SHEM) loss function to {address the issue of overfitting on hard examples}. In the second stage, we propose a sort-based pseudo-labelling fine-tuning (SPF) strategy for semi-supervised learning to mitigate the distributional differences between source and target domains. For both stages, we employ weak or strong image augmentation {to suit different} cross-domain tasks. Experimental results on benchmark datasets {demonstrate} that the proposed network can {enhance} domain adaptation ability for crater detection under varying scenario distributions.
}


\begin{document}

\section{Introduction}
\begin{figure}[H]

    \includegraphics[width=0.7\textwidth, height=0.6\textwidth]{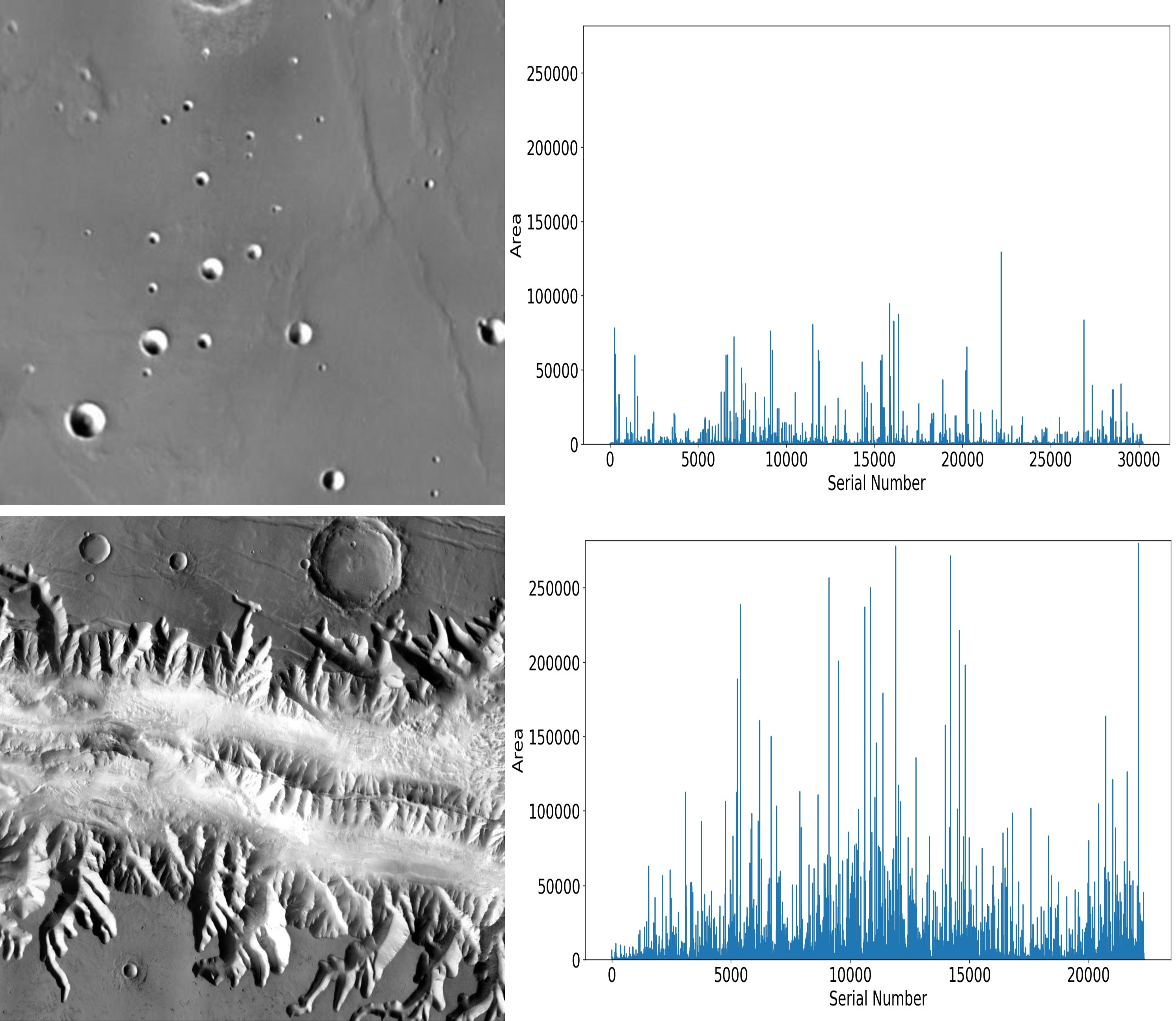}
    \caption{(\textbf{Top}) {One of the} %MDPI: Please use commas to separate thousands for numbers with five or more digits (not for four digits) in the picture. e.g., "10000" should be "10,000"
 samples in the LROC dataset and the distributions of all craters in this dataset. (\textbf{Bottom}) One of the samples in the DACD dataset and the distributions of all craters in this dataset. Compared with the top sample, the bottom one has more background interference. According to the statistical results of these two {datasets, the LROC dataset} contains smaller and more craters than DACD.
       }
       \label{bijiao}
\end{figure}

Delving into the topography and history of extraterrestrial planets (e.g., Mars and the Moon) through the analysis of impact craters is important for humans to explore outer space and broaden their cognition scope. Recent studies on crater detection have been conducted in the field of {computer vision}. However, the exploration of craters on these planets encounters various challenges~\cite{noise_domain_adaptation_1,2023unseen_domain,plus2}. For example, craters on different planets can exhibit significant differences in scale characteristics, density distributions, and background interference. Figure \ref{bijiao} shows two sample images {from} the LROC (Lunar Reconnaissance {Orbiter} Camera) dataset~\cite{lunar} and the DACD (Domain Adaptive Crater Detection) dataset~\cite{DACD}, {along with} the distributions of craters {in terms of} scale variations on these two datasets. Overall, the LROC dataset contains smaller and more craters than the DACD dataset. Some images in the DACD dataset {show} severe background weathering, {which significantly interferes with} crater detection. Hence, detection models trained on known crater datasets are {difficult} to directly apply to new planets due to the domain gap. The development of a robust model with good generalisation abilities under significantly varying scenario distributions remains a great challenge. 

{Training} a crater detection model that can adapt to the terrain of unknown exoplanets {is challenging}. Recent works~\cite{DACD,DEM_carter,MDCD,yang2021craterdanet,crater1} have attempted to {address this issue}. For example, the studies in~\cite{DACD,MDCD,crater1} employ attention mechanisms or use dilated convolution to learn better feature representations. The work in~\cite{yang2021craterdanet} trains the model using generated data. However, due to the lack of sufficient real extraterrestrial data, existing generative models are not always effective. The studies in~\cite{DACD,crater1} develop progressive sampling strategies based on the subspace along the geodesic on the Grassmann manifold or Hough transforms to {enhance} the performance of crater detection. These methods assume that the cross-domain data distributions between source and target domains are similar, but in reality {they vary significantly from one planet to another.} To better learn the distribution of unknown target domains, {some studies employ} semi-supervised pseudo-labelling~\cite{banjiandu1,banjiandu2}, consistent strategy training~\cite{cosis3}, and weakly supervised attention mechanisms~\cite{2023detr_weakly}. In addition, some {studies}~\cite{LRMloss_first,LRM_2020_OIM} {attempt} to mine hard samples by {using} a penalty function {to enhance} the {model's robustness}. However, due to the significant domain gap, they do not generalise well to unknown distributions.

% The existing method based on multi-scale model~\cite{SpatialFPN,jstars2,wang2023yolov7} in the large benchmark data sets to achieve the good effect but migrated to craters detection field, will appear as the model depth deepening, small target characteristics will disappear.

To improve {the} generalisation performance, in this paper, we propose a two-stage adaptive network (TAN) for semi-supervised cross-domain crater detection under varying scenario distributions. Our network is built on the YOLOv5 detector where a series of strategies are employed to enhance the cross-domain generalisation ability. In the first stage, we propose an attention-based scale-adaptive fusion (ASAF) strategy to handle objects with significant scale variances. We propose a smoothing hard example mining (SHEM) loss function to solve the {issue of model overfitting on challenging examples}. In the second stage, we propose a sort-based pseudo-labelling fine-tuning (SPF) strategy for semi-supervised learning to mitigate the distributional differences {between the} source and target domains. For both stages, we {employ} weak or strong image augmentation to {suit different} cross-domain tasks. Experimental results demonstrate the {network's strong} domain adaptation ability for crater detection under varying scenario distributions.

The main contributions of this paper are {summarised} as follows:
\begin{enumerate}[leftmargin=2.1em,labelsep=0.5mm]
  \item[(1)] We present %MDPI: We added the left bracket, please confirm. 
 an attention-based scale-adaptive fusion (ASAF) strategy. This strategy improves the model's scale adaptability, which is helpful {for detecting} craters with significant scale variations.
  
  \item[(2)] We introduce a smoothing hard example mining (SHEM) loss function {to address the issue that the model trained on the source domain is biased towards hard examples in the source domain, leading to poor generalisation to the target domain.}
  
  \item[(3)] We design a sort-based pseudo-labelling fine-tuning (SPF) strategy. {In SPF, we use the trained model from the source domain to generate high-quality pseudo-labels for the target domain.} Subsequently, we fine-tune the model to make it adapt well to the target domain.
  
  \item[(4)] We adopt weak or strong image augmentation to suit different cross-domain tasks by {taking into account} the distribution characteristics of extraterrestrial planets.
\end{enumerate}
%%%%%%%%%%%%%%%%%%%%%%%%%%%%%%%%%%%%%%%%%%
\section{Related Work}\label{Related Work}
{Over the past several years, object detection based on deep learning~\cite{ren2015faster,Lin_2017_CVPR,jstars1,jstars2} has made great progress. Crater detection is also gradually gaining attention~\cite{DACD,DEM_carter,MDCD,yang2021craterdanet,crater1}.} However, there are still several issues that have not yet been fully resolved, including the extreme imbalance of crater sizes, domain adaptation, and the absence of  sufficiently labelled datasets.

\subsection{Object Detection }
For object detection, mainstream methods are based on feature pyramids and feature fusion. After the feature pyramid network (FPN)~\cite{Lin_2017_CVPR} was {developed}, there have been many variants~\cite{SpatialFPN}. For example, the work in~\cite{shen2022hsgm} directly assigns shallow and deep features to different scales. After that, many methods such as top-down integration~\cite{Lin_2017_CVPR} and two-way integration~\cite{panet} were {developed}. The classical approach~\cite{ren2015faster} continuously reduces the feature map from top to bottom while fusing shallow features. Some subsequent studies have investigated bidirectional fusion features~\cite{SpatialFPN}, or even more complex circular recursive fusion features~\cite{diguifpn}. However, most of {the methods mentioned} extract deep convolutional features {through} continuous {subsampling, which is not ideal for detecting small objects}.

A global receptive field can {serve as an effective} equaliser between large and small targets. {The study in~\cite{ASPP} examines the various receptive fields of images, while~\cite{kongdongconv} employs hollow convolution to broaden the perceptual field, where attention is used to gather more contextual information.} Some studies replace all the core convolutions of the backbone or neck network with attention modules~\cite{cbam,biformer}. Although these methods are good solutions to specific problems, shallow knowledge is still forgotten as the network layer deepens, and the information of small targets cannot be effectively propagated to deeper {layers} in the network.

Considering multiple-scale objects, image augmentation methods such as stitching and blurring~\cite{mutilscale-augmentation1} are used to balance the performance {of} large and small target detection. Because most of the hard-to-detect targets are small, researchers have conducted some data {augmentation} studies on small targets~\cite{mutilscale-augmentation1,mutilscale-augmentation4}. However, these augmentation methods are not suitable for cross-domain object detection, {where uneven data distributions may be present in the source and target domains.}

In view of the above problems, in this paper we propose the ASAF strategy to fuse shallow information and {improve} the detection ability of small objects. Furthermore, we {employ different} levels of data enhancement strategies (e.g., weak or strong image augmentation) to {suit distinct} cross-domain {tasks by taking into account} the distribution characteristics of extraterrestrial planets.

\subsection{Domain-Adaptive Object Detection } Domain-adaptive object detection has achieved {significant} results~\cite{domain1,DACD,plus1,plus2}. In the field of crater detection, the target domain {involves} unknown extraterrestrial terrains and interference scenarios. Hence, researchers have made attempts in the directions of sample adaptation and model adaptation, most of which are based on the {methods} of adversarial learning or mutual supervised learning.

Some sample {adaptation} methods apply alignment strategies to address the differences between the source and target domains. For example, Ref. %MDPI: We revised the citation format, please confirm.
~\cite{yangben4} {attempts} to align spatial distributions {through} intrinsic knowledge mining and relational knowledge constraints. Ref.~\cite{yangben6} uses  adversarial training and pseudo-labelling to adapt to the target domain. In this way, the model gradually learns the target domain, and the data in the target domain {guide} the classifiers to achieve a cross-domain effect. However, these methods require a {thorough} understanding of the similarities and differences {between the} target and source domains. If the target domain and its spatial distribution are unknown, these methods may not \mbox{be effective.}

Some model {adaptation} methods are based on the idea of cross-domain-adaptive modules. The design of adaptive modules allows unsupervised adaptation of the model to the target domain. Some studies have {developed} modules to {help assess} the consistency between domains~\cite{moxing2} for {improved} alignment. There are also modules based on knowledge-mining strategies~\cite{moxing4}, by which the images are re-modelled and the network is guided to distinguish the similarities and differences between the domains.

Inspired by the above works, in this paper, we propose the SPF strategy to make the model trained on the source domain learn the distribution of the target domain after fine-tuning and obtain high-quality pseudo-labels.

% {which invovle re-modeling the images and guiding the network to distinguish the similarities and differences between the domains.}

% Inspired by the above works, {this paper introduces the SPF strategy. The SPF strategy aims to make the model trained on the source domain to learn the distribution of the target domain through fine-tuning and acquire high-quality pseudo-labels.}

\subsection{Object Detection with a Scarcity of Labelled Data}

\textls[-15]{Currently, there is no high-quality benchmark dataset for crater detection, and the {quality of datasets} is often uneven. {To the best of our knowledge}, only {a few} crater \mbox{datasets~\cite{DEM_carter,Crater_v1_Robbins,yang2021craterdanet}}} have been {made available} and there {is currently} no {extensive} benchmark dataset {similar to} COCO or PASCAL VOC. Because deep learning is data-hungry, generative \mbox{learning~\cite{diffusion-zongshu,yang2021craterdanet,GAN-mage}} has gained attention {for acquiring} training data. There are also some works {that use text} to generate images~\cite{GAN-texttoimage2023}. {Due} to the scarcity of labelled data, {it is currently challenging to synthesise real crater data.}

{Recent studies}~\cite{cosis3,banjiandu1,yangben6} have {employed} semi-supervised algorithms to {address} the {limited} labelled data {through} consistency-based learning or pseudo-labelling. Consistency-based learning algorithms use two networks to learn the consistency of output results for the same unlabelled images under {various} perturbations. As a result, information from unlabelled data is fully utilised. Pseudo-labelling algorithms~\cite{banjiandu1,yangben6} use models pre-trained on labelled data to {generate} pseudo-labels {for} unlabelled data {during} model training.

In {case of limited} labelled data, some studies~\cite{loss1,LRM_2020_OIM} {tackle} this {issue} by increasing the penalty {for erroneous samples}, or by {employing} hard example mining strategies~\cite{hardminingloss1}. However, these studies do not {address} the cross-domain problem. When applied to cross-domain crater detection, the model is more biased {towards challenging} examples in the source domain, which fails to generalise well to the target domain with simple distributions.

{Due to the great disparity of craters on different exoplanets, we need to develop some strategies to make the model adaptive to varying scenario distributions.} Based on the above works, in this paper, we propose the SHEM loss function. {It aims to identify challenging} examples in the source domain with complex distributions, {while facilitating the model's adaptation} to the target domain with {simpler} distributions. By contrast, if the source domain has relatively simple distributions, we only adopt the common hard example mining loss to make our model {adaptive} to the target domain with complex distributions. 

\section{Proposed Method}\label{Proposed Method}
\begin{figure}[H]

    \includegraphics[width=0.95\textwidth,height=0.49\textwidth]{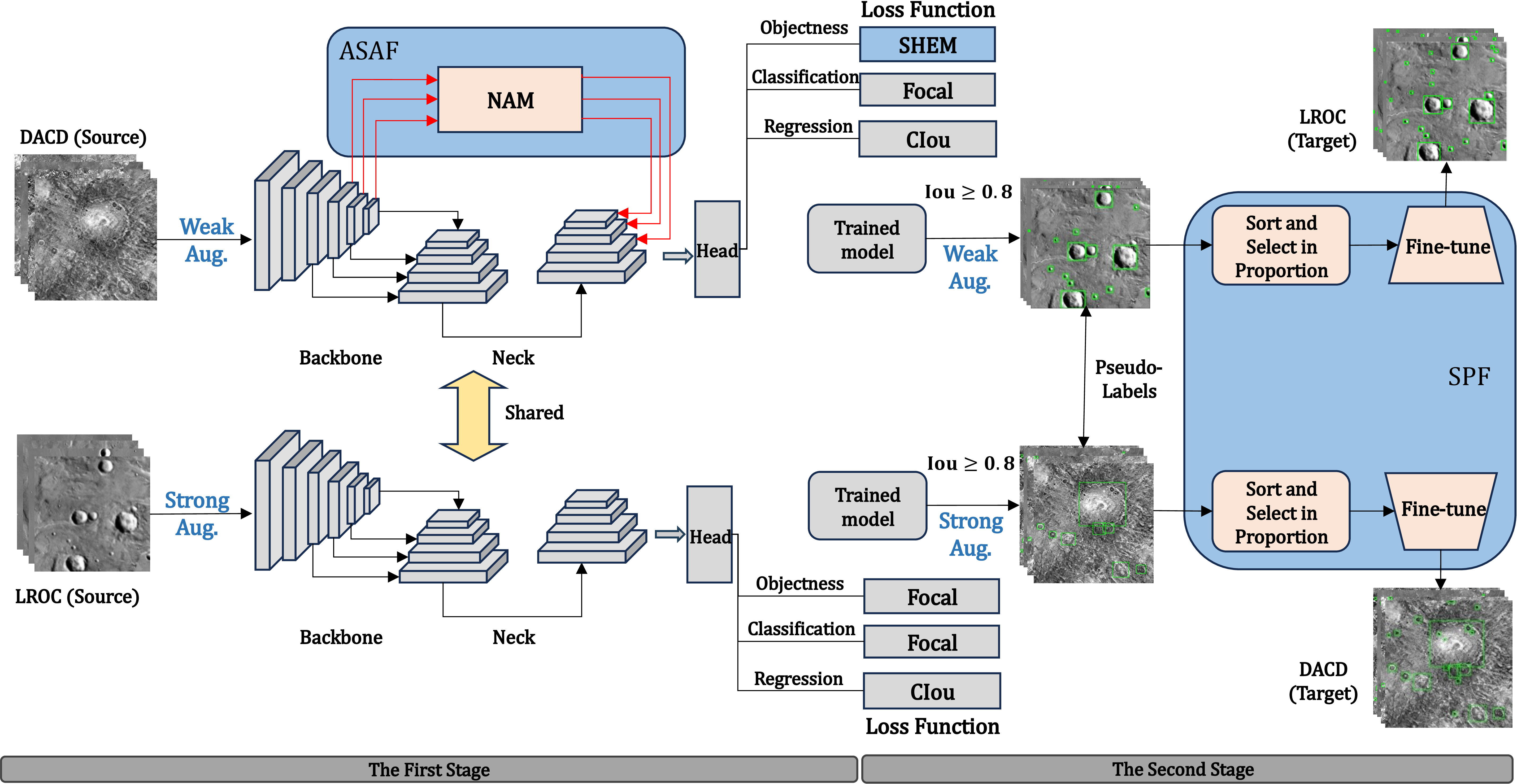}
    \caption{The architecture %MDPI: Figure should be placed close to where it is first mentioned, so we moved it here, please confirm.
 of our proposed two-stage TAN model for cross-domain crater detection. The first stage: (1) ASAF utilises the attention-based NAM (Normalisation-Based Attention Module) to fuse shallow information to improve scale adaptation abilities; (2) the SHEM loss function is used to alleviate the bias of the model. The second stage: The SPF strategy is adopted to sort and select high-quality pseudo-labels {which are used to  fine-tune the model}. In these two stages, we adopt weak or strong image augmentation to match different cross-domain tasks. {The new components are highlighted in blue font or regions.}
    }
    \label{fig1}
\end{figure}

Figure \ref{fig1} shows the proposed framework TAN, which mainly {consists of} the following four new {components:} the attention-based scale-adaptive fusion (ASAF) strategy, the smoothing hard example mining (SHEM) loss functions, the sorting-based pseudo-labelling fine-tuning (SPF), and the data augmentation strategy.

Specifically, Figure \ref{fig1} illustrates two types of cross-domain tasks. The {upper section of} this figure {illustrates} cross-domain detection {using} the DACD dataset as the source domain and the LROC dataset as the target domain. This is a detection task from the complex to the simple domain. Hence, we adopt weak augmentation, ASAF, and SHEM to train our model in the source domain, followed by fine-tuning to learn the distribution of the target domain.

The bottom part {of} this figure {illustrates} cross-domain detection from simple to complex. That is, we {use the LROC dataset as the source domain and the DACD dataset as the target domain.} In this case, we rely on strong augmentation and the focal function (without ASAF) to mine {challenging} examples in the source domain. {This approach} facilitates subsequent learning in the target domain with complex distributions. 

{The motivations for the use of different first-stage training processes are as follows. For the cross-domain problem, craters in the source and target domains can have different scenario distributions (complex or simple). If the source domain is complex, we use a weak augmentation method to prevent the model over-learning the complex features. We also use the ASAF strategy to mitigate the extreme scale variations in craters (too big or small). In addition, we use the SHEM to mitigate the model bias towards hard examples and achieve better generalisation ability. On the contrary, if the source domain is simple, we use the strong augmentation to enhance the robustness of our model, and the ASAF and SHEM are accordingly no longer needed.}

% It is a detection task from the complex to the simple domain. Hence, we adopt weak augmentation, NAM  (Normalization-based Attention Module) and SHEM to train our model in the source domain, followed by fine-tuning to learn the distribution of the target domain.

% The bottom part in this figure is for cross-domain detection from simple to complex. That is, we employ the LROC and DACD as the source and target domain, respectively. In this case, we rely on strong augmentation and the focal function (without NAM) to mine hard examples in the source domain, which facilitates the subsequent learning in the target domain for complex distributions. 

\subsection{Attention-Based Scale-Adaptive Fusion (ASAF)}
As the convolutional layers deepen, shallow and subtle target features {are more likely} to be overlooked. In addition, due to limited labelled data and potential interference, models tend to be biased towards the training data. Therefore, integrating the initial feature maps and emphasising informative features can mitigate this bias and enhance the \mbox{model's robustness}. 

\begin{figure}[H]

    \includegraphics[width=0.89\textwidth]{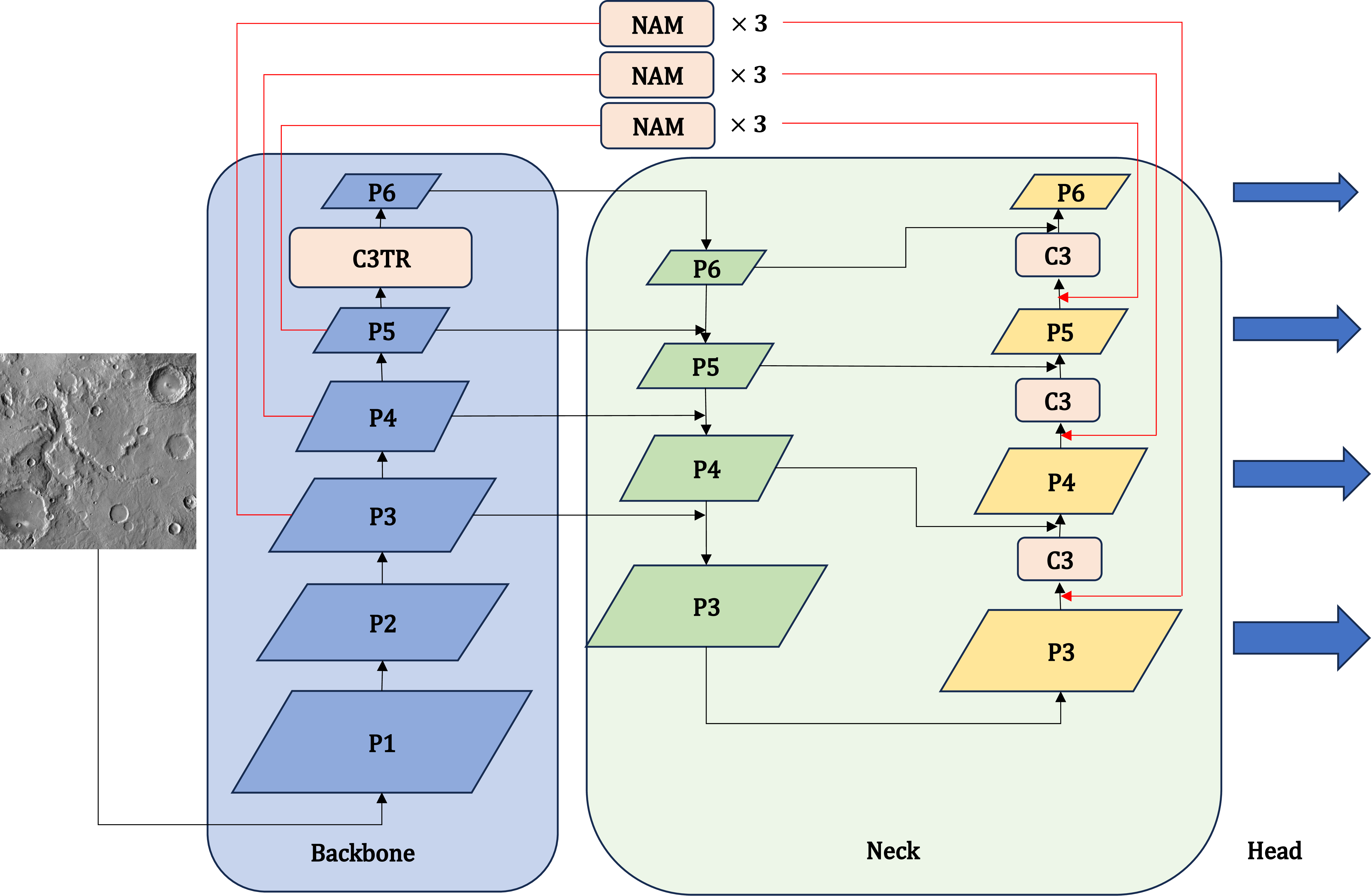}
    \caption{Illustration of our model architecture with the ASAF strategy. To prevent the model from losing crucial information, we incorporate C3TR (C3 + Transformer) into the backbone. We pass the shallow feature maps of different stages to NAM to obtain shallow attention-based feature maps. We also inject multiple C3 modules {into} the neck {to detect} large targets.}
    \label{ASAF}
\end{figure}

Motivated by these observations, we introduce an ASAF strategy to integrate shallow information into the deeper layers. {Specifically, our model architecture is based on the YOLOv5~\cite{YOLOv5}, which consists of three main components: backbone, neck, and head. The core modules of the backbone and neck are C3, including three Conv blocks and a bottleneck block. Additionally, the backbone incorporates the focus structure, while the neck utilises the FPN~\cite{Lin_2017_CVPR} with the PAN (Path Aggregation Network)~\cite{panet} architecture. We pass the shallow feature maps of different stages to the NAM (Normalisation-Based Attention Module)~\cite{nam}, and then, fuse the feature maps P3, P4, and P5 in the backbone and the neck (refer to Figure \ref{ASAF}). Here, P denotes the feature maps of various sizes acquired from different layers of the network in the YOLOv5 architecture. Furthermore, to extract global long-range features, we incorporate C3TR (C3 + Transformer block~\cite{transfomer}) into the backbone by replacing a series of bottlenecks in the C3 with one Transformer block~\cite{transfomer}.} In addition, we inject multiple C3 modules {into} the neck after fusing shallow attention-based feature maps. These C3 modules collectively contribute to improved performance for detecting large targets.

\vspace{-6pt}

\begin{figure}[H]

    \includegraphics[width=0.9\textwidth]{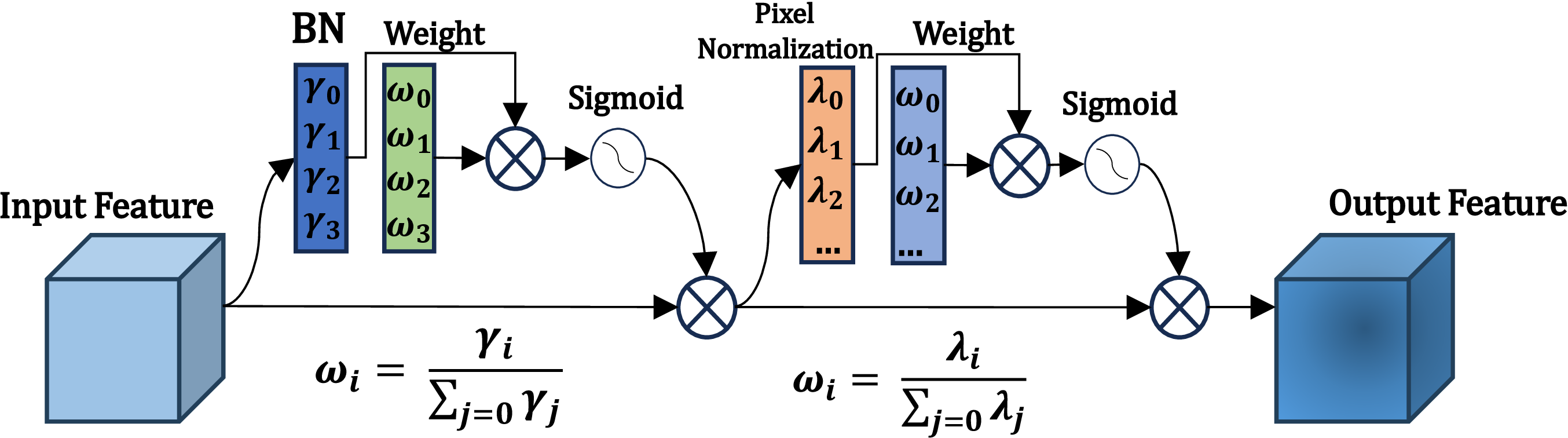}
    \caption{The architecture of NAM~\cite{nam}.}
    \label{f}
\end{figure}

The details of NAMs~\cite{nam} as attention mechanism modules are shown in Figure \ref{f}. Each NAM contains the CBAM (Convolutional Block Attention Module)~\cite{cbam} with some {modified} channel and spatial attention sub-modules. To be specific, the channel attention sub-module uses the scale factor in batch normalisation (\emph{BN}), which measures the variance in the channels and indicates their significance. This can be formulated as follows: %MDPI: Please make sure that the format of equations is consistent in all of text.

\begin{equation}
B_{out} = BN(B_{in}) = \gamma \frac{B_{in}-\mu_{B}}{\sqrt{\sigma_{B}^{2} + \varepsilon} }+ \beta
\end{equation}
where $\mu_{B}$ and $\sigma_{B}$ are the mean and standard deviation of the mini-batch $B$, respectively; $\gamma$ and $\beta$ are trainable affine transformation parameters (i.e., scale and shift). Assume that F1 is the input feature and $M_c$ is the output feature; they are given in the following equations:
\begin{equation}
M_c = sigmoid(W_{\gamma}(BN(F_1)))
\end{equation}
where $M_c$ represents the output feature, $\gamma$ is the scaling factor for each channel, and the weights are obtained as $W_{\gamma} = \gamma_i/\sum_{j=0}{\gamma_j}$. Similarly, for the spatial attention sub-module, the formula for the spatial attention mechanism is given below:
\begin{equation}
M_s = sigmoid(W_{\lambda}(BN_s(F_2)))
\end{equation}
where $\lambda$ is the scaling factor of BN (i.e., pixel normalisation in Figure \ref{f}), the weights are $W_{\lambda} = \lambda_i/\sum_{j=0}{\lambda_j}$, and the output is denoted as $M_s$.

\vspace{-4pt}

\subsection{Smoothing Hard Example Mining (SHEM)}

 The original YOLOv5 utilised the binary cross-entropy (BCE) loss function for classification and objectness. However, we found that the focal loss~\cite{focalloss} {outperforms} BCE for binary crater detection due to its {superior} ability to balance positive and negative examples. The focal loss function is computed as follows:
\begin{equation}
L_{fl}=\left\{
\begin{aligned}
&-(1-\hat{q})^\lambda log(\hat{q}), \quad y=1 \\
&-\hat{q}^\lambda log(1-\hat{q}),  \enspace\;\quad y=0
\end{aligned}
\right.
\end{equation}

\begin{equation}
q=\left\{
\begin{aligned}
&\hat{q}, \quad\quad\enspace\; y=1 \\
&1-\hat{q},  \quad\text{otherwise}
\end{aligned}
\right.
\end{equation}
where $y$ {represents} the true label, $q$ {represents} the predicted probability, $\lambda$ {represents} the focal parameter, and $\alpha$ is the correction parameter. The above equations can be {summarised} as the following equation:
\begin{equation}
% Focal(q)=-\alpha (1-q)^{\lambda}log(q)
L_{fl} = -(1-q)^{\lambda}log(q)
\end{equation}

To address challenging target detection where significant background interference and complex distributions are present, we introduce a loss function to mine hard examples. First, we employ the balance factor $\xi$ to enhance the focal loss function, known as the equilibrium focal loss (BFL)~\cite{LRMloss_first}, defined as 
\begin{equation} \label{bfl}
L_{BFL} = \xi L_{fl}, \quad \xi > 1\\
\end{equation}

\begin{figure}[H]

   \includegraphics[width=0.9\textwidth]{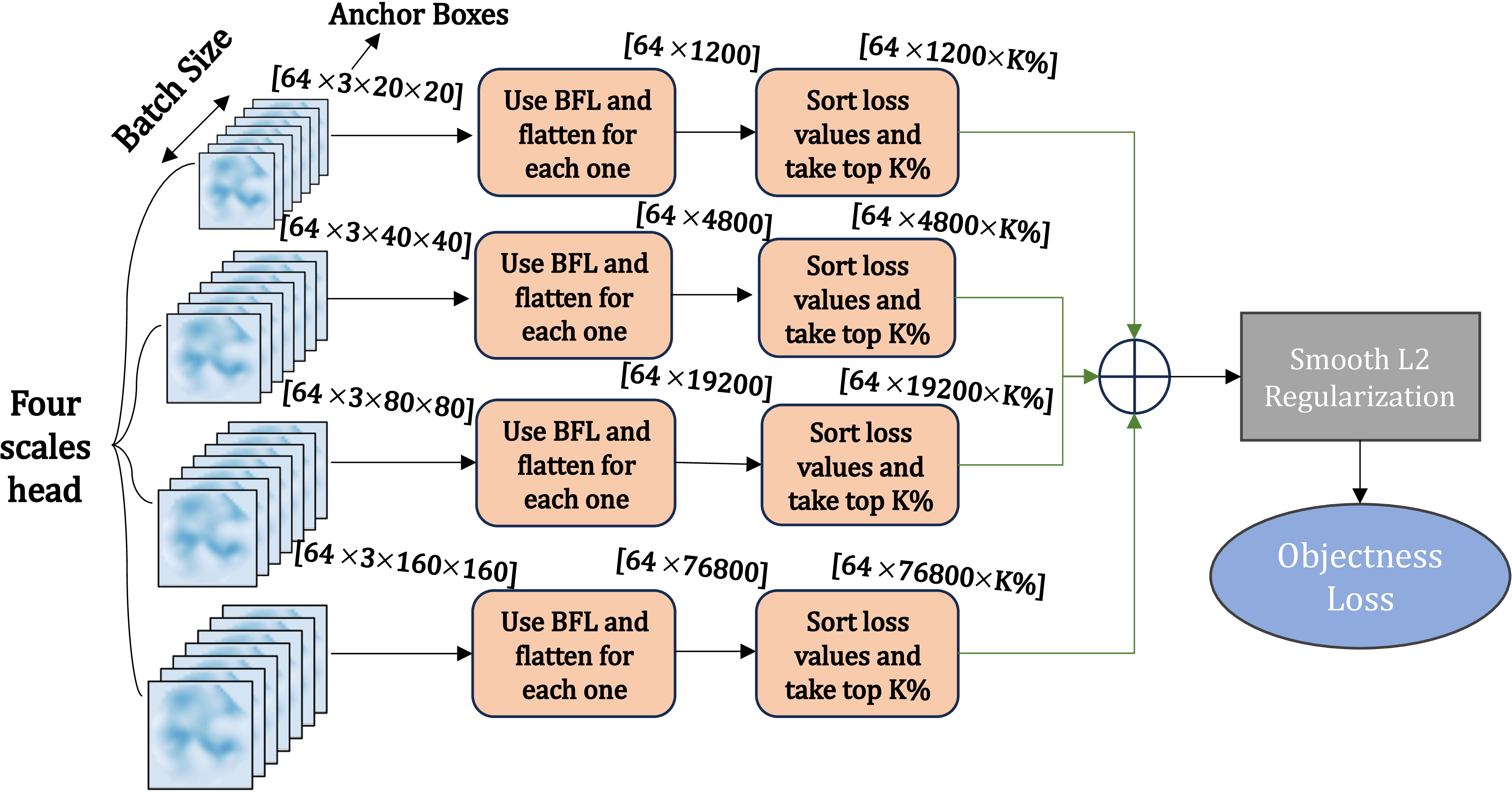}
    \caption{The overall %MDPI: Please use commas to separate thousands for numbers with five or more digits (not for four digits) in the picture. e.g., "10000" should be "10,000"
 flow of the SHEM loss function. We {calculate} BFLs for the feature maps at four scales, {each with a distinct distribution of loss values.} Then, we select the top K\% of {loss values that} have been sorted for the feature maps. Subsequently, we average and weigh the loss values at different scales to obtain the Loss Rank Function (LRM), followed by L2 regularisation to obtain the objectness loss.}
    \label{fig3}
\end{figure}

Based on Equation (\ref{bfl}), we {calculate} BFLs for the feature maps at {various scales, each with a distinct distribution of loss values.} Then, we select the {top K\% of loss values that} have been sorted as the loss values for the feature maps. Subsequently, we average and weigh the loss values at different scales to obtain the Loss Rank Function (LRM). Considering the substantial variations in crater scales, we extended the final head in the YOLOv5 model to four scales. While the LRM~\cite{LRMloss_first} loss function enables the model to learn numerous hard examples, its emphasis on hard examples may lead to overfitting on the source domain dataset. Therefore, we introduce the SHEM loss function to {calculate} the objectness loss. The computation framework of the SHEM loss function is depicted in Figure \ref{fig3}, and can be formulated as follows:
\begin{equation}
L_{SHEM} = L_{LRM} + \lambda ||w||^2
\end{equation}
where $w$ denotes a weight vector, and $\lambda$ denotes a regularisation parameter (we set its value to $ 5\times10^{-9}$). Here, we utilise L2 regularisation to ensure the model gradually learns hard examples and mitigate the risk of overfitting. As shown in Figure \ref{fig1}, we use the three losses, i.e., SHEM, focal, and complete intersection over union (CIoU) losses to train our first-stage model for cross-domain detection (from complex to simple).

\vspace{-2pt}

\subsection{Sorting-Based Pseudo-Labelling Fine-Tuning (SPF)}
 
The gap between the source and target domains for crater detection can be significant, especially in the case that the target domain is entirely unknown to the source domain and lacks label information. Therefore, it is impractical to use the conventional learning model to obtain the feature distribution of the target domain. To address this issue, we employ pseudo-labelling to fine-tune the {model, as} detailed below. 

The source domain dataset with labels is denoted as $D_1 = \{x_i, y_i\}^{N_1}$, where $N_1$ is the total number of samples, $x_i$ {denotes} the image sample, and $y_i$ {denotes} the corresponding sample labels. The target domain dataset without labels is denoted as $D_2 = \{x_i\}^{N_2}$, where $N_2$ is the total number of samples in the target domain. After obtaining the trained detection model in the source domain (denoted as $M_1$), as introduced {previously}, we apply $M_1$ to the unlabelled target domain $D_2$ to obtain the pseudo-labels $\hat{y}_i$ for the image  $x_i$, i.e.,
\begin{equation}
\{\hat{y}_i \}= M_1(D_2), \quad IoU \ge 0.8, \quad i=1,...,N_2\\
\end{equation}

To fully learn the distribution of the target domain with high-quality pseudo-labels, we sort images in descending order {based on} the number of detected objects in each sample and {choose a subset} of them as our fine-tuning training set. The formula is as follows:

% \begin{equation}
% N_s = \alpha \times N_1 \\
% \end{equation}
\begin{equation}
\{\tilde{y}_j\} = select_h(sort(\{\hat{y}_i \})
\end{equation}

Here, $\tilde{y}_j$ is the pseudo-labels for the $j$th image after sorting and selection, and $h$ is the selected proportion, defined as
\begin{equation}
h=\frac{N_1 \times \alpha}{N_2} \leq 0.3
\end{equation}
where $\alpha$ is a control parameter, {which is used to balance the number of pseudo-labelled data used for fine-tuning and the source data. The balance is achieved by setting a relatively small $\alpha$ value if the amount of source data is much larger than that of the target data, and a large $\alpha$ value if the amount of source data is much smaller than that of the target data.}

\textls[-15]{Subsequently, we fine-tune the pre-trained model $M_1$, which is obtained in the first stage, on $\{\tilde{y}_j\} $ as well as the corresponding images for 2--3 rounds by freezing the backbone parameters (the first 10 layers in the model). The calculation process is illustrated in Figure \ref{fig1}.}

\subsection{Data Augmentation Strategy}

A challenge arises when the model trained on the source domain fails to adapt to the target domain due to a lack of understanding of the target domain's distributions. Moreover, in the new domain, the model may encounter interference from unknown environments, and is influenced by factors such as different signal-to-interference ratios and background complexity. These domain gaps make the model hard to transfer between different domains, resulting in inaccurate detection results.

To address this issue, we propose a domain-adaptive data augmentation strategy for cross-domain detection that distinguishes between complex and simple data distributions. As shown in Figure \ref{fig1}, models trained on the source domain with less background interference tend to be less robust. {To address this issue}, we employ strong data augmentation techniques (e.g., affine transformation and random mosaic) to increase the complexity {of the} training data. Conversely, for the source domain data with complex distributions, we {only apply} minor data augmentation techniques (e.g., simple flipping and stitching) to ensure that the model learns from smoother data. In this way, the model becomes better adapted to the smoothed domain for cross-domain detection.

\section{Experiments}

In this section,  we first introduce the datasets used for our experiments, and then, introduce the used evaluation metrics and our implementation details. Next, we compare our method with the state of the art (SOTA) and perform ablation experiments. After visualising the detection results, we   provide  some discussions on our method.

\subsection{Datasets}
We use the following three datasets for our experiments.

\textbf{\emph{DACD Dataset. %MDPI: Please confirm if the bolds and italics should be retained.
}}
The DACD~\cite{DACD} (Domain Adaptive Crater Detection) dataset contains two scenes of craters: the day craters and the night craters
on Mars. It consists of \mbox{1000 images} from $90^{\circ}$S to $90^{\circ}$N
latitude and $180^{\circ}$W to $180^{\circ}$E longitude, leading to more than
20,000 craters in total. The size of these craters is about
5 pixels to 300 pixels. The scenario of the dataset contains a lot of noisy interference.

\textbf{\emph{LROC Dataset.}}
To validate the effectiveness of our model for solving the cross-domain problem, we choose a public crater dataset~\cite{lunar}, which is obtained by the Lunar Reconnaissance Orbiter Camera (LROC). It consists of 866 images of size 416 × 416 pixels captured on %EE: Please check meaning retained
the Moon. The size of most of the craters is between 3 pixels and 200 pixels, and the craters are relatively densely distributed.

\textbf{\emph{DOTA Dataset.}}
In addition, we demonstrate the generalisation ability of our model on the large public remote sensing dataset DOTA~\cite{dota} (A Large-scale Dataset for Object Detection in Aerial Images). It consists of 2806 aerial images from different sensors and platforms. There are 15 categories of images (about 4000-by-4000 pixels) which exhibit a variety of scales, orientations, and shapes.

\subsection{Evaluation Metrics}  %用的什么指标，页数不够写详细，页数够了缩略写
We use recall, precision and mean average precision (mAP) to evaluate our model; these are defined as follows.

The {quality} of a single detection box is {determined by} the IoU value, which is the intersection ratio between the predicted and actual detection frames. We use recall and precision to evaluate the performance of the network model by setting the IoU threshold to 0.5 by default, i.e., 
\begin{equation}
Recall = \frac{TP}{TP+FN}
\end{equation}
\begin{equation}
Precision = \frac{TP}{TP+FP}
\end{equation}

{When} the IoU of the detection frame is greater than or equal to 0.5, it is {classified} as a true positive (\emph{TP}%MDPI: Please unify the format of variables in the formula and text (italic or not).
); otherwise, the detected crater is {classified} as a false positive (\emph{FP}). \emph{FN} {represents} the number of false negatives, and \emph{TN} is the number of negative samples that are correctly identified. 

In addition to the above metrics, we use mAP to {calculate} the mean value of AP for all classes, where AP is the area under the precision--recall curve. In our experiments, we {calculate} two kinds of mAPs: (1) mAP@.5 by setting IoU to 0.5; (2) mAP@.5:.95 by varying IoU from 0.5 to 0.95 with an interval of 0.05 and {averaging the results.}

\subsection{Implementation Details} %写一些超参数，输入大小 优化器， 学习率等等,还有训练轮之类的
 In our experiments, we use the YOLOv5~\cite{YOLOv5} as our baseline. For dataset splitting, we split the source dataset into training and validation sets in a ratio of 8:2. We use Pytorch as our deep learning framework, and we resize the images to $640 \times 640$ pixels during training. We adopt SGD~\cite{hu2023bag} as our optimiser and use momentum gradient descent as our learning strategy with {an} initial learning rate of 0.01, 3 rounds of iteration once, and {a} weight decay of 0.0005. All the experiments are run on a 3.70 GHz and 18-processor Intel Xeon W-2255 CPU, and an Nvidia GeForce RTX 3080Ti GPU with 12 G %EE: AE - please check units
 memory.

\subsection{Comparison with State-of-the-Art Methods} %有几个数据集你就比较几个段落
In this section, some comparisons with other models are conducted on the three datasets. We use MMDetection~\cite{mmdet} as a tool. According to {their} working principle, some backbone networks are introduced for object detection in some models.

\subsubsection{Comparisons on DACD Dataset and Cross-Domain from DACD to LROC}

Some SOTA detection methods are selected in Table \ref{table1} for comparison with our model. Compared with some two-stage models (e.g., Faster R-CNN~\cite{ren2015faster} and Libra R-CNN~\cite{libra_rcnn}), our model {demonstrates superior precision} {but} slightly {inferior recall performance}. Compared with single-stage detection models (e.g., YOLOv5~\cite{YOLOv5} and YOLOv7~\cite{wang2023yolov7}), our model achieves better detection performance in both precision and recall. In Table \ref{table1}, the PVT has only a slightly higher recall, but the precision is much lower than our model. Finally, for cross-domain detection (from the DACD dataset to the LROC dataset), our model slightly improves the baseline YOLOv5 in precision, but achieves a significantly better result in recall (about 24.04\%) than the baseline. 

\begin{table}[H]
\tablesize{\fontsize{9}{9}\selectfont}

    \caption{Performance comparison %MDPI: We remove the label {tab2l}. Table should be placed close to where it is first mentioned, so we placed it here. Please confirm.
 on the DACD dataset and cross-domain detection from DACD to LROC.}

    \setlength{\cellWidtha}{\textwidth/5-2\tabcolsep--0.6in}
\setlength{\cellWidthb}{\textwidth/5-2\tabcolsep-0.15in}
\setlength{\cellWidthc}{\textwidth/5-2\tabcolsep-0.15in}
\setlength{\cellWidthd}{\textwidth/5-2\tabcolsep-0.15in}
\setlength{\cellWidthe}{\textwidth/5-2\tabcolsep-0.15in}
\scalebox{1}[1]{\begin{tabularx}{\textwidth}{
>{\PreserveBackslash\centering}m{\cellWidtha}
>{\PreserveBackslash\centering}m{\cellWidthb}
>{\PreserveBackslash\centering}m{\cellWidthc}
>{\PreserveBackslash\centering}m{\cellWidthd}
>{\PreserveBackslash\centering}m{\cellWidthe}}

\toprule
    \multirow{2.7}{*}{\textbf{Model}} & \multicolumn{2}{c}{\textbf{DACD Dataset}} &  \multicolumn{2}{c}{\textbf{Migrate to LROC Dataset}} \\ 
    % \midrule
    \cmidrule{2-5}
                &\textbf{Precision (\%)}&  \textbf{Recall (\%)} & \textbf{Precision (\%)} &  \textbf{Recall (\%)} \\ 
    \midrule
         DAFaster~\cite{DAFaster} & $55.15$ & $50.91$& $35.25$ & $30.18$\\
         EPM~\cite{EPM} & $55.69$ & $48.09$& $35.53$ & $29.06$\\
         MeGA~\cite{mega}& $60.22$ & $52.81$& $40.24$ & $34.57$\\
         PDAN~\cite{DACD} & $62.31$ & $54.83$& $46.83$ & $38.77$\\
         Faster R-CNN~\cite{ren2015faster}& $ 74.76 $&$79.89$ & $ 45.57 $&$ 47.06$\\
         PAFPN~\cite{panet}& $79.87 $&$ 74.52 $& $40.05$&$ 41.58 $ \\
         DCN~\cite{dcn} & $ 75.17 $&$ 80.14\ $& $ 47.46 $&$ 45.53 $ \\
         SABL~\cite{sabl}& $ 73.82 $&$ 79.05 $& $ 47.31 $&$ 45.32 $ \\
         Deformable-DETR~\cite{deformable_detr}& $ 79.71 $& $\textbf{92.20 %MDPI: Please confirm if the bold should be retained.
}$& $ 41.19 $& $53.09$\\
         Libra R-CNN~\cite{libra_rcnn}& $ 71.18 $& $79.57$& $ 47.13 $& $49.88$\\
         PVT~\cite{pvt}& $ 77.58 $&$ 89.77 $& $ 54.99 $&$ \textbf{62.81} $ \\
         ATSS~\cite{atss}& $ 80.84 $&$\underline{91.67 %MDPI: Please confirm if the underline should be retained.
} $& $ 56.89 $&$ 62.73 $ \\
         YOLOv5~\cite{YOLOv5} & $ \underline{81.05} $& $77.51$& $ \underline{91.51} $& $38.74$\\
         YOLOv7~\cite{wang2023yolov7} & $ 80.70 $& $77.91$& $ 67.42 $& $43.56$\\
         Ours & $ \textbf{83.64} $& $78.50$& $  \textbf{92.78} $& $\underline{62.78}$\\
         
    \bottomrule
    \end{tabularx}}
    \label{table1}
\end{table}

These results show that our proposed ASAF can {achieve} good generic detection performance by combining shallow information without cross-domain detection. After migrating to the LROC dataset, our model achieves the best results in terms of both precision and recall, demonstrating the effectiveness of SPF used in the second stage.

\subsubsection{Comparisons on LROC Dataset and Cross-Domain from LROC to DACD}
{As shown in Table \ref{table2}, the proposed model achieves the best performance in terms of precision. Due to the shallow information propagation using an attention-based scale-adaptive fusion strategy, our model can accurately detect small craters in the samples from the LROC dataset. When the model trained on the LROC dataset is tested on the DACD dataset, our model improves the baseline YOLOv5~\cite{YOLOv5} by 6.02\% in terms of recall. This is because the use of SPF and the data augmentation strategy can adapt to the distribution of the target domain. The recall of our model is worse than that of PVT~\cite{pvt} and ATSS~\cite{atss}. The reasons are as follows. The DACD dataset consists of high-resolution images which contain crater objects with unbalanced distributions, while LROC consists of images containing dense small crater objects. As ATSS is dedicated to addressing the unbalanced sample problems for object detection, it generally achieves good results on the DACD dataset. Due to the use of the convolution-free pyramid attention mechanism, PVT has a natural advantage for the dense small object detection task (see the results of recall on LROC), and still maintains excellent performance on the DACD dataset for the cross-domain problem.}

\begin{table}[H]
\tablesize{\fontsize{9}{9}\selectfont}
    \caption{Performance comparison %MDPI: We remove the \label{tab3}, please confirm.
 on the LROC dataset and cross-domain detection from LROC to DACD.}
    
    \setlength{\cellWidtha}{\textwidth/5-2\tabcolsep--0.6in}
\setlength{\cellWidthb}{\textwidth/5-2\tabcolsep-0.15in}
\setlength{\cellWidthc}{\textwidth/5-2\tabcolsep-0.15in}
\setlength{\cellWidthd}{\textwidth/5-2\tabcolsep-0.15in}
\setlength{\cellWidthe}{\textwidth/5-2\tabcolsep-0.15in}
\scalebox{1}[1]{\begin{tabularx}{\textwidth}{
>{\PreserveBackslash\centering}m{\cellWidtha}
>{\PreserveBackslash\centering}m{\cellWidthb}
>{\PreserveBackslash\centering}m{\cellWidthc}
>{\PreserveBackslash\centering}m{\cellWidthd}
>{\PreserveBackslash\centering}m{\cellWidthe}}

\toprule
    \multirow{2.5}{*}{\textbf{Model}} & \multicolumn{2}{c}{\textbf{LROC Dataset}} &  \multicolumn{2}{c}{\textbf{Migrate to DACD Dataset}} \\ 
    % \midrule
    \cmidrule{2-5}
                &\textbf{Precision (\%)}&  \textbf{Recall (\%)} & \textbf{Precision (\%)} &  \textbf{Recall (\%)} \\ 
    \midrule
         DAFaster~\cite{DAFaster} & $50.12$ & $45.81$& $34.66$ & $36.43$\\
         EPM~\cite{EPM} & $49.23$ & $44.01$& $35.02$ & $36.77$\\
         MeGA~\cite{mega}& $53.48$ & $49.01$& $41.39$ & $42.98$\\
         PDAN~\cite{DACD} & $54.33$ & $50.83$& $45.41$ & $45.77$\\
         Faster R-CNN~\cite{ren2015faster}& $ 56.52 $&$ 57.75$& $ 49.95 $&$ 55.55$ \\
         PAFPN~\cite{panet}& $57.50 $&$ 59.00 $& $50.93 $&$ 56.16 $ \\
         DCN~\cite{dcn} & $ 56.62$&$ 57.53 $& $ 49.32 $&$ 54.21 $ \\
         SABL~\cite{sabl}& $ 57.60 $&$ 58.75 $&$ 49.85 $&$ 54.77 $ \\
         Deformable-DETR~\cite{deformable_detr}& $ 81.55 $& $86.65$& $ 55.11 $& $59.35$\\
         Libra R-CNN~\cite{libra_rcnn}&  $ 56.54 $& $ 58.66 $& $ 50.21 $& $56.65$\\
         PVT~\cite{pvt}& $ 87.90 $&$ \textbf{92.36} $& $ 55.04 $&$ \underline{72.49} $ \\
         ATSS~\cite{atss}& $ 86.31 $&$ \underline{91.19} $&$ 61.95 $&$ \textbf{78.34} $ \\
         YOLOv5~\cite{YOLOv5} & $  92.06$& $84.45$& $ 64.73$& $52.26$\\
         YOLOv7~\cite{wang2023yolov7} & $ \underline{93.20} $& $89.60$ & $ \underline{69.80} $& $52.10$\\
         Ours & $ \textbf{93.94}$& $89.78$& $\textbf{69.82}$& $58.28$\\
    \bottomrule
    \end{tabularx}}
    \label{table2}
\end{table}

\subsubsection{Comparisons on the DOTA Dataset}

In this section, we use a large remote sensing object detection dataset, DOTA~\cite{dota}, to {evaluate} the model's performance on the generic detection task. The DOTA dataset contains many small targets and has a wide variety of scenarios. We follow the training and testing {settings} defined in~\cite{dota} to evaluate the models on the DOTA dataset. We show the results in Table \ref{table3}, where we have listed the AP value for each object category. The abbreviations of the categories are described as follows: BD---baseball field, GTF---runway, SV---small vehicle, LV---large vehicle, TC---tennis court, BC---basketball court, SC---storage tank, SBF---soccer field, RA---roundabout, SP---swimming pool, HC---helicopter. As can be seen from Table \ref{table3}, our model shows better mAP performance {compared to} other {methods, with the exception of} YOLOv7. However, as shown in Tables \ref{table1} and \ref{table2}, YOLOv7~\cite{wang2023yolov7} {exhibits poorer} performance than {our model} for cross-domain crater detection. There are also some models~\cite{atss,pvt} that {perform} well when the DACD dataset is used as the source domain, but {show} poor generalisation and robustness on the DOTA dataset  {because they struggle} to adapt to complex changes. This proves that our model can guarantee no performance degradation {in} the generic remote image object detection task, in addition {to demonstrating} good generalisation abilities for cross-domain crater detection.
\startlandscape
%\text{~}\vspace{-18pt}
\begin{table}[H]

    \caption{Performance comparison  on the DOTA dataset.}
   
\setlength{\cellWidtha}{\textwidth/17-2\tabcolsep--1in}
\setlength{\cellWidthb}{\textwidth/17-2\tabcolsep-0in}
\setlength{\cellWidthc}{\textwidth/17-2\tabcolsep-0.1in}
\setlength{\cellWidthd}{\textwidth/17-2\tabcolsep-0.1in}
\setlength{\cellWidthe}{\textwidth/17-2\tabcolsep-0.1in}
\setlength{\cellWidthf}{\textwidth/17-2\tabcolsep-0.1in}
\setlength{\cellWidthg}{\textwidth/17-2\tabcolsep-0.1in}
\setlength{\cellWidthh}{\textwidth/17-2\tabcolsep-0.1in}
\setlength{\cellWidthi}{\textwidth/17-2\tabcolsep-0.1in}
\setlength{\cellWidthj}{\textwidth/17-2\tabcolsep-0.1in}
\setlength{\cellWidthk}{\textwidth/17-2\tabcolsep-0.1in}
\setlength{\cellWidthl}{\textwidth/17-2\tabcolsep-0.1in}
\setlength{\cellWidthm}{\textwidth/17-2\tabcolsep-0.1in}
\setlength{\cellWidthn}{\textwidth/17-2\tabcolsep-0.1in}
\setlength{\cellWidtho}{\textwidth/17-2\tabcolsep-0.1in}
\setlength{\cellWidthp}{\textwidth/17-2\tabcolsep-0.1in}
\setlength{\cellWidthq}{\textwidth/17-2\tabcolsep--0.3in}
\scalebox{1}[1]{\begin{tabularx}{\textwidth}{
>{\PreserveBackslash\centering}m{\cellWidtha}
>{\PreserveBackslash\centering}m{\cellWidthb}
>{\PreserveBackslash\centering}m{\cellWidthc}
>{\PreserveBackslash\centering}m{\cellWidthd}
>{\PreserveBackslash\centering}m{\cellWidthe}
>{\PreserveBackslash\centering}m{\cellWidthf}
>{\PreserveBackslash\centering}m{\cellWidthg}
>{\PreserveBackslash\centering}m{\cellWidthh}
>{\PreserveBackslash\centering}m{\cellWidthi}
>{\PreserveBackslash\centering}m{\cellWidthj}
>{\PreserveBackslash\centering}m{\cellWidthk}
>{\PreserveBackslash\centering}m{\cellWidthl}
>{\PreserveBackslash\centering}m{\cellWidthm}
>{\PreserveBackslash\centering}m{\cellWidthn}
>{\PreserveBackslash\centering}m{\cellWidtho}
>{\PreserveBackslash\centering}m{\cellWidthp}
>{\PreserveBackslash\centering}m{\cellWidthq}} 

\toprule
\textbf{Model}&\textbf{SV}&\textbf{LV}&\textbf{Plane}&\textbf{ST}&\textbf{Ship}&\textbf{Harbour}&\textbf{GTF}&\textbf{SBF}&\textbf{TC}&\textbf{SP}&\textbf{BD}&\textbf{RA}&\textbf{BC}&\textbf{Bridge}&\textbf{HC}&\textbf{mAP@.5 (\%)}\\

\midrule
Faster R-CNN~\cite{ren2015faster}&26.59&53.11&61.99&32.38&36.90&39.96&41.65&40.14&78.53&20.56&37.29&28.84&37.77&17.50&26.04&61.44\\
\midrule
DCN~\cite{dcn}&26.97&55.08&62.38&34.00&38.75&41.96&43.05&\underline{42.33 %MDPI: Please confirm if the underline should be retained.
}&79.72&22.58&39.00&29.89&42.23&20.17&25.31&63.36\\
\midrule

SABL~\cite{sabl}&27.43&56.66&64.47&33.62&38.88&41.82&40.75&39.40&81.60&22.47&41.02&30.94&40.29&18.31&28.29&62.00\\
\midrule
ATSS~\cite{atss}&23.96&46.20&56.93&31.89&34.15&30.87&26.68&25.35&74.22&21.19&34.32&26.21&30.74&12.26&23.26&56.81\\
\midrule
PVT~\cite{pvt}&21.22&43.56&57.95&28.90&33.21&34.19&33.34&23.43&75.20&21.10&34.54&26.71&30.76&14.52&27.44&57.67\\
\midrule

Libra R-CNN~\cite{libra_rcnn}&27.19&53.89&62.27&33.46&39.32&38.74&\underline{43.18}&40.99&78.60&20.99&38.86&29.91&38.35&18.97&24.76&62.17\\
\midrule
Deformable-DETR~\cite{deformable_detr}&15.34&33.50&56.09&24.51&22.78&28.60&22.67&27.49&70.88&21.57&32.53&24.28&24.94&17.48&23.54&54.28\\
\midrule
YOLOv5~\cite{YOLOv5}&\underline{40.53}&\underline{61.66}&68.21&41.74&\underline{61.17}&47.36&37.38&41.53&82.74&\underline{27.91}&42.03&\underline{33.70}&\underline{47.97}&\textbf{23.09 %MDPI: Please confirm if the bold should be retained.
}&\textbf{35.82}&71.69\\
\midrule
YOLOv7~\cite{wang2023yolov7}&\textbf{43.80}&\textbf{66.70}&\textbf{70.70}&\textbf{48.60}&\textbf{64.80}&\textbf{52.20}&\textbf{47.71}&\textbf{52.59}&\textbf{87.28}&\textbf{28.02}&\underline{45.21}&30.11&\textbf{59.63}&\underline{23.07}&29.98&\textbf{74.40}\\
\midrule
Ours&40.33&60.75&\underline{68.25}&\underline{41.87}&61.16&\underline{47.84}&36.60&40.05&\underline{83.02}&26.79&\textbf{45.66}&\textbf{33.88}&46.44&22.62&\underline{32.83}&\underline{71.80}\\
\bottomrule
		\end{tabularx}}
		\label{table3}
\end{table}
\finishlandscape
%\text{~}\vspace{-18pt}

\subsection{Ablation Studies}  %根据自己模块设计
The comparison results presented in Tables \ref{table1}--\ref{table3} demonstrate that the proposed TAN is superior to many state-of-the-art methods. In the following section, we comprehensively analyse TAN from three aspects to explore its superiority. For this purpose, we perform ablation experiments by using DACD as the source domain and LROC as the target domain.

\begin{figure}[H]

    \subfloat{
        \includegraphics[width=0.76\textwidth,height=0.48\textwidth]{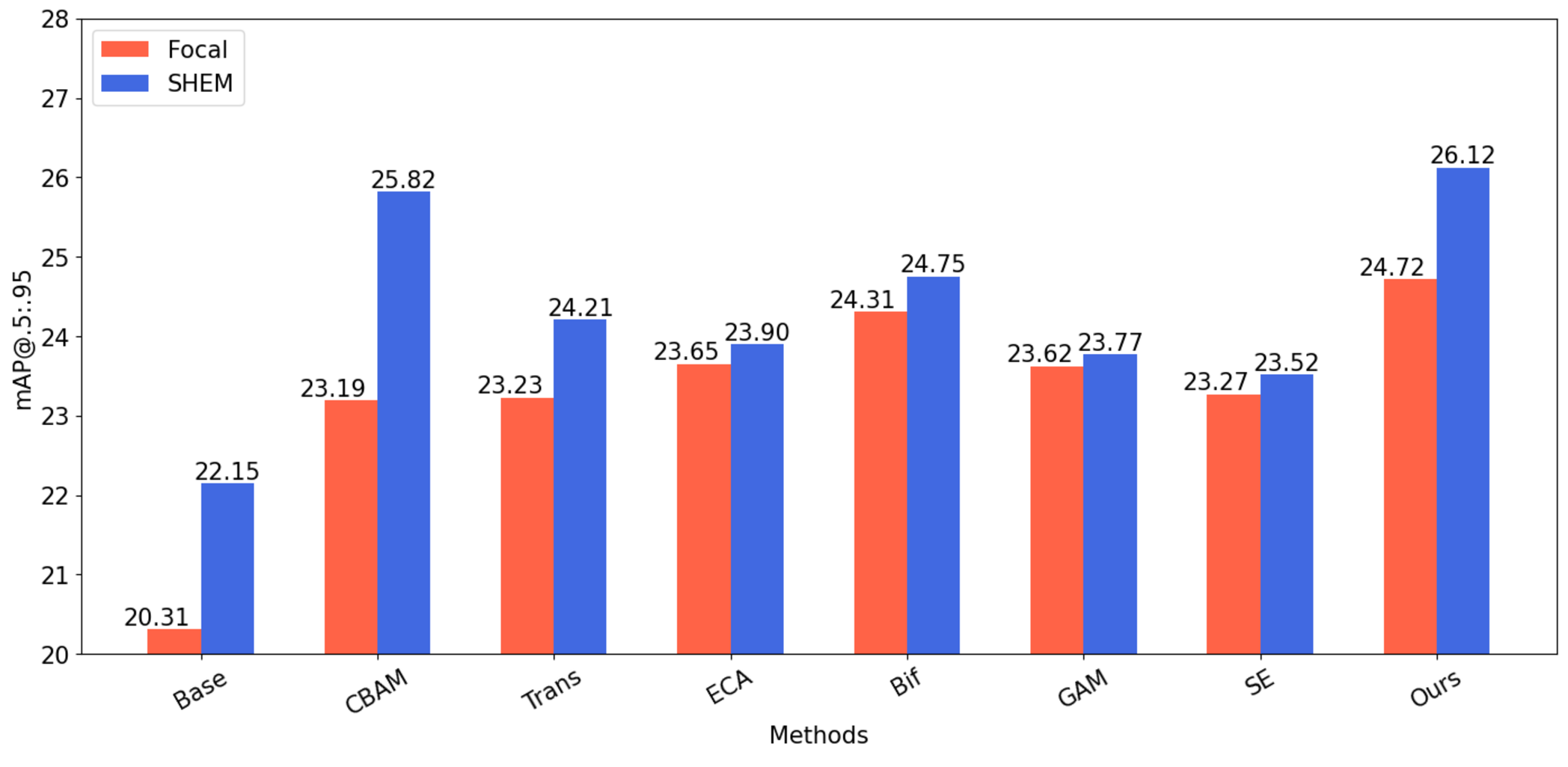}
        \label{fig:subfigeee}
    }\\%\vspace{-0.5cm}
    \subfloat{
         \includegraphics[width=0.76\textwidth,height=0.48\textwidth]{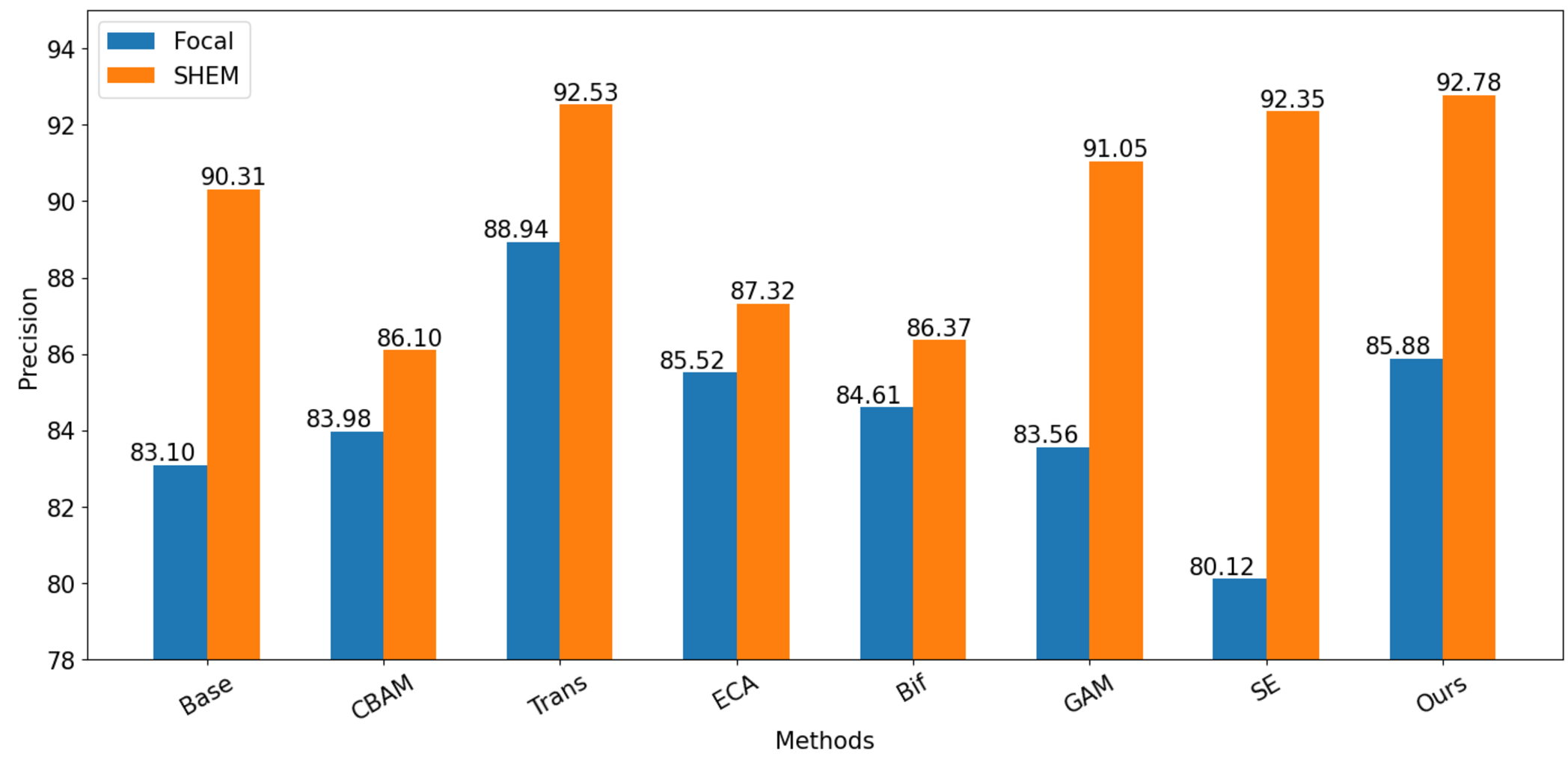}
        \label{fig:subfigaaa}
    }
    % \vspace{-0.1cm}
    \caption{Ablation experiment %MDPI: Figure should be placed close to where it is first mentioned, so we placed it here, please confirm.
 on different attention mechanism modules and loss functions for cross-domain crater detection from DACD to LROC.}
    \label{mokuai}
\end{figure}

(1) {Effects of replacing NAM in the model by using different attention mechanism modules with an SHEM or a focal loss function.} To this end, we do not use data augmentation strategies and SPF, and define YOLOv5 as the basic scheme, which uses the focal loss {for} objectness (without any attention mechanisms). Based on YOLOv5, we evaluate the effects of different attention mechanisms and loss functions on the precision and mAP@.5:.95, {as illustrated} in Figure \ref{mokuai}. {Here, Trans stands for the Transformer block~\cite{transfomer}, Bif stands for the Vision Transformer with Bi-Level Routing Attention block~\cite{biformer}, SE stands for the SE block in Squeeze-and-Excitation Networks~\cite{se}, CBAM stands for the Convolutional Block Attention Module~\cite{cbam}, ECA stands for the Efficient Channel Attention block~\cite{ecanet}, and GAM stands for the Global Attention Mechanism block~\cite{gam}.} As show in Figure \ref{mokuai}, the adopted SHEM outperforms the focal loss with respect to both evaluation indicators. The use of different attention mechanisms can generally improve the performance {of the} basic YOLOv5. Among them models, our adopted NAM~\cite{nam} achieves the best performance. The above results show that our proposed method, using NAM coupled with the SHEM loss function, has good generalisation ability for cross-domain crater detection.

\begin{figure}[H]

    \vspace{-0.2cm}
    \hspace{-1.2cm}\includegraphics[width=1\textwidth,height=0.6\textwidth]{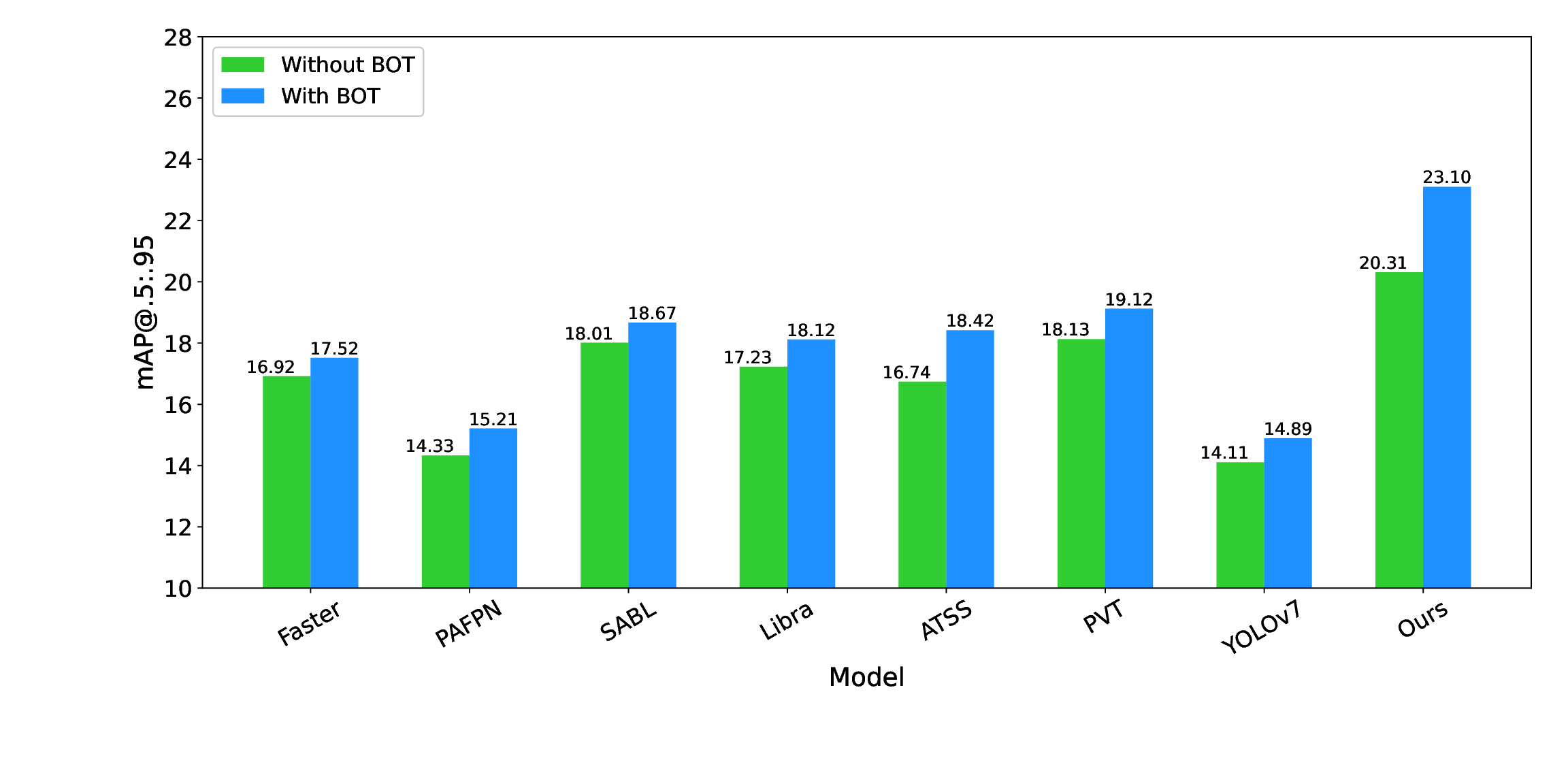}
    \vspace{-0.6cm}\caption{Ablation experiment on BOT (data augmentation and SPF)  for cross-domain crater detection from DACD to  LROC, where the Faster refers to Faster R-CNN~\cite{ren2015faster} and Libra refers to Libra R-CNN~\cite{libra_rcnn}.}
    \label{bot}
\end{figure}

(2) Effects of data augmentation and SPF. The two strategies in our model complement each other, and we refer to it as BOT (bag of tricks) for cross-domain crater detection experiments from DACD to LROC. To validate their effectiveness, we do not use ASAF and SHEM in all the compared models. {Instead}, we compare the results using mAP@.5:.95 as the evaluation indicator. As shown in Figure \ref{bot}, the use of BOT improves the mAP@.5:.95 for all the compared models. In particular, our model {improves} the mAP@.5:.95 value by 2.79 when BOT is used, demonstrating its effectiveness for cross-domain crater detection. {In order to verify the effectiveness of the two strategies individually, we conduct experiments by only using SPF and the resulting mAP@.5:.95 is 21.98\%. By comparing the results obtained using BOT (i.e., 23.10\%) and without using BOT (i.e., 20.31\%), we can see the efficacy of both data augmentation and SPF.}

(3) Effects of different combinations of our proposed components.
{To this end, we conduct} ablation experiments on different combinations of the proposed ASAF, SHEM, and BOT. As can be seen from Table \ref{table4}, each component helps to improve the recall and mAP@.5:.95. When {all the} three components are  used, our model achieves the best results, demonstrating the effectiveness of these components.

\begin{table}[H]

    \caption{Ablation experiments on different combinations of our proposed strategies for cross-domain crater detection from DACD to LROC.}
    
   \setlength{\cellWidtha}{\textwidth/6-2\tabcolsep-0.3in}
\setlength{\cellWidthb}{\textwidth/6-2\tabcolsep-0.1in}
\setlength{\cellWidthc}{\textwidth/6-2\tabcolsep-0.1in}
\setlength{\cellWidthd}{\textwidth/6-2\tabcolsep-0.1in}
\setlength{\cellWidthe}{\textwidth/6-2\tabcolsep-0.1in}
\setlength{\cellWidthf}{\textwidth/6-2\tabcolsep--0.7in}
\scalebox{1}[1]{\begin{tabularx}{\textwidth}{
>{\PreserveBackslash\centering}m{\cellWidtha}
>{\PreserveBackslash\centering}m{\cellWidthb}
>{\PreserveBackslash\centering}m{\cellWidthc}
>{\PreserveBackslash\centering}m{\cellWidthd}
>{\PreserveBackslash\centering}m{\cellWidthe}
>{\PreserveBackslash\centering}m{\cellWidthf}}

\toprule

     ~&\textbf{ASAF}& \textbf{SHEM} & \textbf{BOT} & \textbf{Recall (\%)} & \textbf{mAP@.5:.95 (\%)} \\
\midrule
         1&\ding{55} & \ding{55}  & \ding{55}&38.74 &20.31\\

         2 &\ding{55} & \Checkmark  & \ding{55}&42.88 &22.15\\

         3 &\ding{55} &  \ding{55}  &\Checkmark&43.40 &23.10\\

         4 &\Checkmark&  \ding{55}  &\ding{55}&52.85 &24.72\\

         5 &\Checkmark &  \ding{55}  &\Checkmark&55.63 &25.22\\

         6 &\ding{55} &  \Checkmark  &\Checkmark&56.33 &25.46\\

         7 &\Checkmark &  \Checkmark  &\ding{55}&\underline{58.63} &\underline{26.12}\\

         8 &\Checkmark & \Checkmark &\Checkmark&\textbf{62.78} &\textbf{27.36}\\
\midrule
    \end{tabularx}}
    \label{table4}
\end{table}

\vspace{-6pt}
\unskip

\subsection{Visualisation} 
\begin{figure}[H]
    
    \subfloat[\centering]{
        \includegraphics[width=0.98\textwidth]{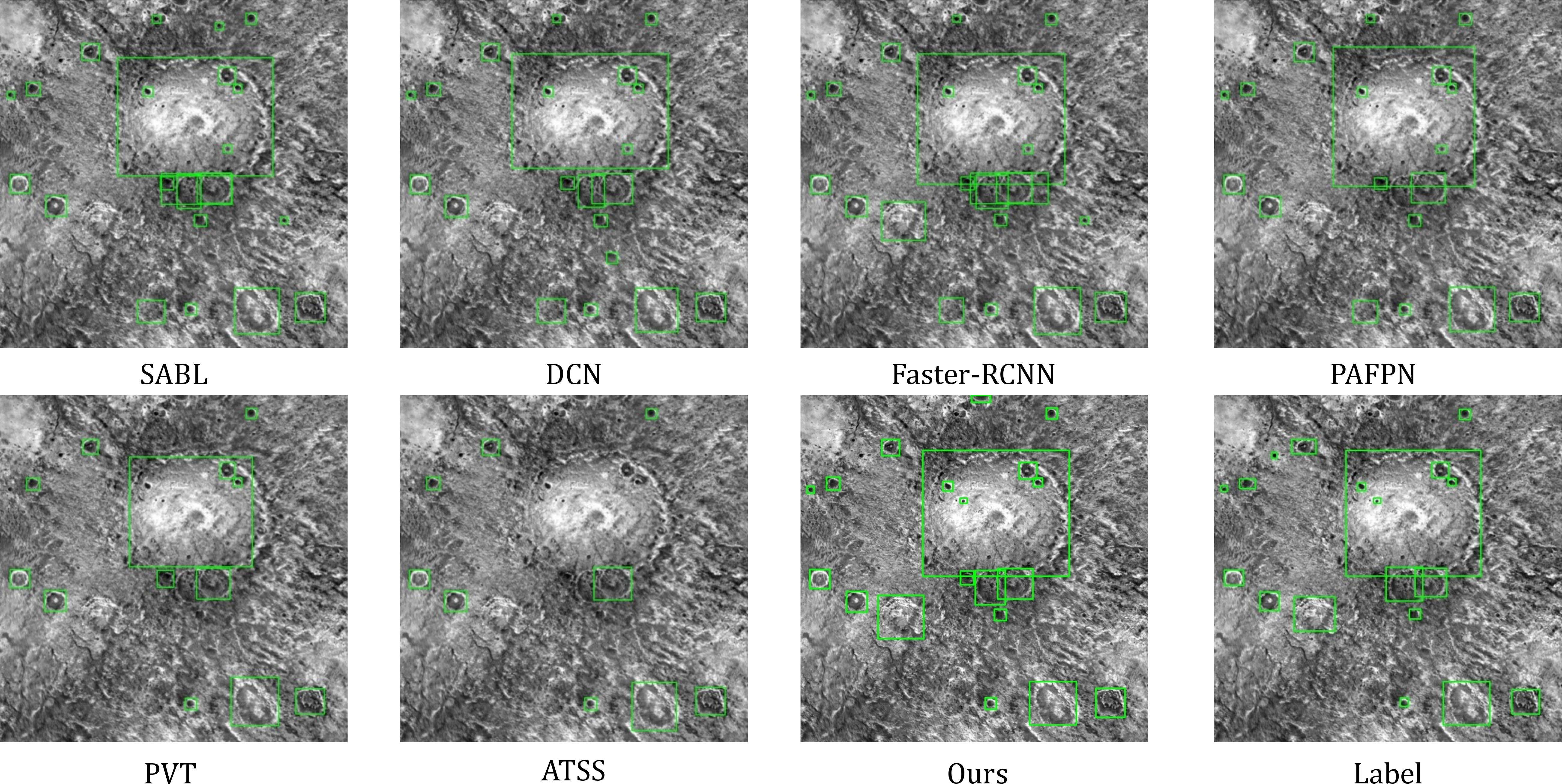}
        \label{fig:subfigeeeee}
    }\\
    \subfloat[\centering]{
        \includegraphics[width=0.98\textwidth]{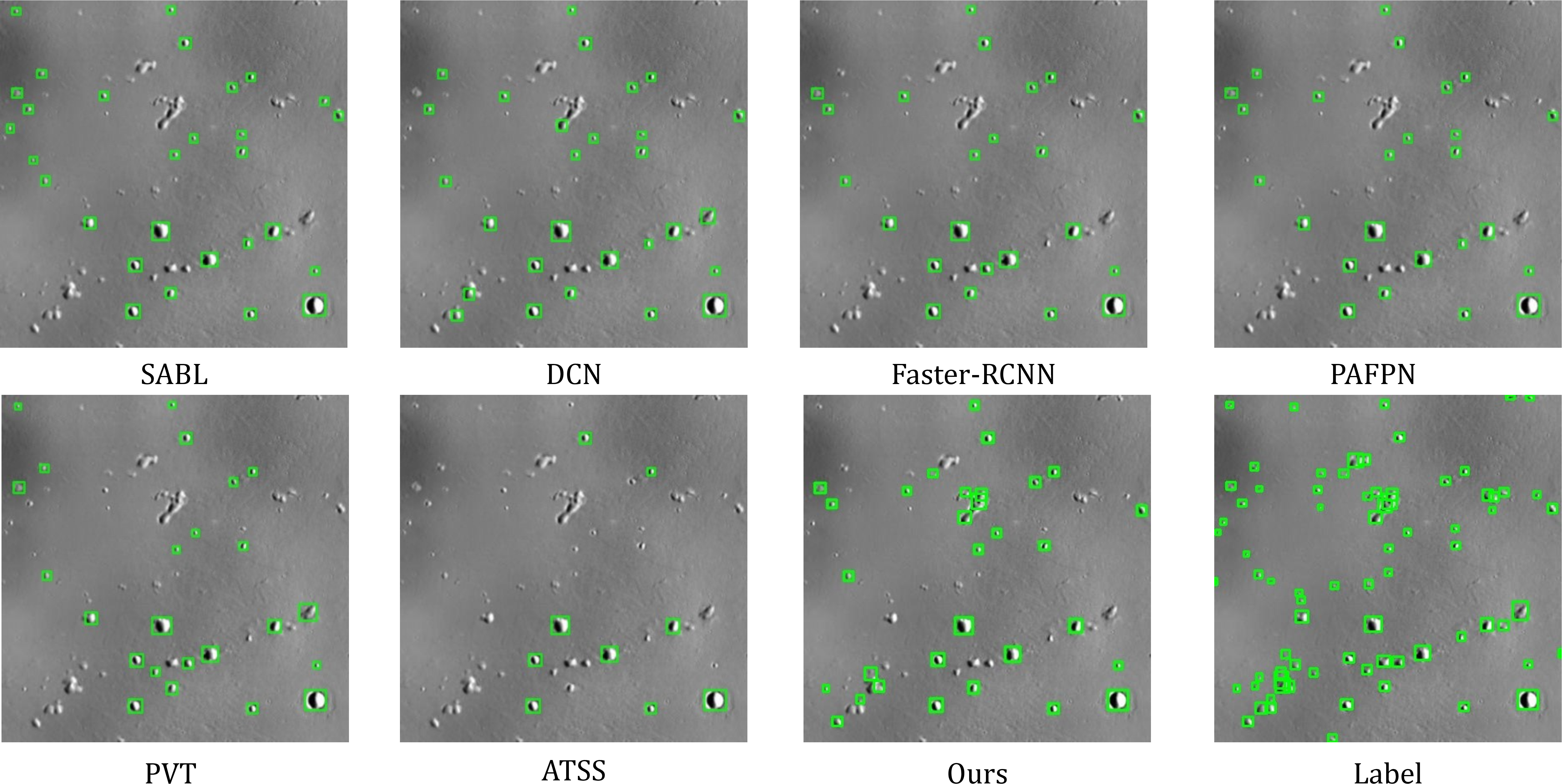}
        \label{fig:subfigaaaaaa}
    }
    \caption{Visualisation of %MDPI: Figure should be placed close to where it is first mentioned, so we placed it here, please confirm.
 the detection results of different models on the LROC dataset as well as the cross-domain detection results. (\textbf{a}). Visualisation of %MDPI: We moved the subfigure explanations into the figure caption. Please confirm.
 the detection results of the proposed method and the mainstream methods on the DACD dataset. (\textbf{b}). Visualisation of the detection results of the proposed method and the mainstream methods for cross-domain detection from DACD to LROC.}
    \label{fig:mainesfs}
\end{figure}

We visualise the detection results of different models on the DACD dataset. As can be seen in Figure \ref{fig:mainesfs}a, our model {demonstrates} good detection results. ATSS~\cite{atss} misses larger-scale craters, PVT~\cite{pvt} {overlooks} some small craters, and Faster R-CNN~\cite{ren2015faster} detects more craters. For the cross-domain detection from DACD to LROC, our model detects more craters than other models, {while} ATSS misses many small craters, as shown in Figure \ref{fig:mainesfs}b. Thus, our model has better generalisation ability, which {enables it to adapt more effectively} to cross-domain scenarios with different background distributions.

\subsection{Further Discussion }
In future research, we aim to  enhance the model's performance based on lower-quality labelled datasets. Our findings {emphasise} the significant challenge of accurately labelling craters in images. The manual labelling process is inherently complex and prone to errors.
As such, our focus will shift towards techniques {for} weakly supervised target detection with incomplete labels or entirely unsupervised targets. We are committed to addressing the issue {of effectively} identifying more craters in scenarios involving poor data quality or incomplete labels.

In addition, we conduct experiments using different backbone networks in Table \ref{table5}. As can be seen, the results obtained by using the backbone networks with a large number of parameters are not necessarily better. For backbone networks with a small number of parameters, such as ConvNeXt~\cite{convnext} and Ghost~\cite{ghost}, the detection results are pretty good. Therefore, in future {research} we will {investigate} lightweight backbone networks {which can still maintain} high-quality detection results for practical applications.

\begin{table}[H]

   \caption{Results obtained %MDPI: Figure should be placed close to where it is first mentioned, so we left it here, please confirm.
 by using different backbone networks on the DACD dataset.}
   
  \setlength{\cellWidtha}{\textwidth/5-2\tabcolsep--0.8in}
\setlength{\cellWidthb}{\textwidth/5-2\tabcolsep-0.2in}
\setlength{\cellWidthc}{\textwidth/5-2\tabcolsep-0.25in}
\setlength{\cellWidthd}{\textwidth/5-2\tabcolsep-0.1in}
\setlength{\cellWidthe}{\textwidth/5-2\tabcolsep-0.25in}
\scalebox{1}[1]{\begin{tabularx}{\textwidth}{
>{\PreserveBackslash\centering}m{\cellWidtha}
>{\PreserveBackslash\centering}m{\cellWidthb}
>{\PreserveBackslash\centering}m{\cellWidthc}
>{\PreserveBackslash\centering}m{\cellWidthd}
>{\PreserveBackslash\centering}m{\cellWidthe}}

\toprule
  \textbf{Backbone}& \textbf{Parameters} & \textbf{FLOPs} & \textbf{Precision (\%)} & \textbf{Recall (\%)}\\
\midrule
Res2Net~\cite{res2net}  &73.29M& 171.8G & 74.30 & 48.14 \\
\midrule
ResNet-50~\cite{resnet} & 27.56M& 72.2G& 81.48& 76.96\\
\midrule
ResNet-18~\cite{resnet} & 12.35M & 31.6G & \underline{81.71 %MDPI: Please confirm if the underline should be retained.
} & 76.74\\
\midrule
GhostNet~\cite{ghost} & 3.69M&  8.2G & 79.84 &\underline{78.09}\\
\midrule
ConvNeXt~\cite{convnext} &1.59M& 4.1G & 79.39& 76.81\\
\midrule
CSPDarkNet53~\cite{cspnet} (Original) &12.32M& 16.3G &81.05 &77.51\\
\midrule
CSPDarkNet53~\cite{cspnet} (Ours) &14.94M& 18.4G &\textbf{83.64 %MDPI: Please confirm if the bold should be retained.
} &\textbf{78.50}\\
\midrule
		\end{tabularx}}
		\label{table5}
\end{table}

\section{Conclusions}\label{Conclusion} %用过去式
In this study, we propose TAN, a two-stage adaptive network for semi-supervised cross-domain crater detection. TAN is {based} on the YOLOv5 detector, {which incorporates} a series of strategies to {improve} the cross-domain generalisation abilities. In the first stage, we propose an attention-based scale-adaptive fusion strategy to {address the scale variation issue} of crater objects. We also propose a smoothing hard example mining loss function to solve the issue of overfitting on hard examples. In the second stage, we propose a sort-based pseudo-labelling fine-tuning strategy by using the trained model in the first stage to {address} the distributional differences {between the} source and target domains. For both the stages mentioned above, we employ weak or strong image augmentation to {suit} different cross-domain tasks (from complex to simple or vice versa).  

To validate the performance of our network, we conduct ablation experiments on the DACD and LROC datasets, as well as cross-domain experiments. We also validate the generality of our model on a large remote sensing {dataset called} DOTA. All of our ablation experiments demonstrate the strong domain-adaptive ability of our model under varying scenario distributions.

\textbf{Future Work. %MDPI: Please confirm if the bold should be retained.
} We will further improve our model's generalisation ability, {especially} in case of poor data quality or {very limited} labelled data. In addition, we will  {explore} lightweight models {which can maintain} good detection results.

\vspace{6pt}

\authorcontributions{ Methodology, Y.L.; Software, Y.L.; Validation, Y.L.; Formal analysis, Y.Z.; Resources, R.L.; Data curation, C.X.; Writing---original draft, Y.L.; Writing---review \& editing, T.S. and T.G.; Visualization, R.C.; Project administration, T.S. All authors have read and agreed to the published version of the manuscript.%MDPI: For research articles with several authors, a short paragraph specifying their individual contributions must be provided. The following statements should be used ``Conceptualization, X.X. and Y.Y.; methodology, X.X.; software, X.X.; validation, X.X., Y.Y. and Z.Z.; formal analysis, X.X.; investigation, X.X.; resources, X.X.; data curation, X.X.; writing---original draft preparation, X.X.; writing---review and editing, X.X.; visualization, X.X.; supervision, X.X.; project administration, X.X.; funding acquisition, Y.Y. All authors have read and agreed to the published version of the manuscript.'', please turn to the  \href{http://img.mdpi.org/data/contributor-role-instruction.pdf}{CRediT taxonomy} for the term explanation. Authorship must be limited to those who have contributed substantially to the work~reported.
}

\funding{
This work was supported by the  National Natural Science Foundation of China (62371084), the Natural Science Foundation of Chongqing, China (CSTB2022NSCQ-MSX1418), the China Postdoctoral Science Foundation (2022MD723727), the Special Support for Chongqing Postdoctoral Research Project (2022CQBSHTB2041), and the Funding of Institute for Advanced Sciences of Chongqing University of Posts and Telecommunications (E011A2022330).
 %MDPI: Please add: ``This research received no external funding'' or ``This research was funded by NAME OF FUNDER grant number XXX.'' and  and ``The APC was funded by XXX''. Check carefully that the details given are accurate and use the standard spelling of funding agency names at \url{https://search.crossref.org/funding}, any errors may affect your future funding.
}

\dataavailability{The public sources of the data mentioned in this study have been described in the paper. %MDPI: We encourage all authors of articles published in MDPI journals to share their research data. In this section, please provide details regarding where data supporting reported results can be found, including links to publicly archived datasets analyzed or generated during the study. Where no new data were created, or where data is unavailable due to privacy or ethical restrictions, a statement is still required. Suggested Data Availability Statements are available in section ``MDPI Research Data Policies'' at \url{https://www.mdpi.com/ethics}.
}

\conflictsofinterest{The authors declare no conflicts of interest. %MDPI: Declare conflicts of interest or state ``The authors declare no conflicts of interest.'' Authors must identify and declare any personal circumstances or interest that may be perceived as inappropriately influencing the representation or interpretation of reported research results. Any role of the funders in the design of the study; in the collection, analyses or interpretation of data; in the writing of the manuscript; or in the decision to publish the results must be declared in this section. If there is no role, please state ``The funders had no role in the design of the study; in the collection, analyses, or interpretation of data; in the writing of the manuscript; or in the decision to publish the results''.
}

\begin{adjustwidth}{-\extralength}{0cm}

\reftitle{References}

\PublishersNote{}
\end{adjustwidth}
\end{document}